\documentclass[letterpaper,10pt,conference]{ieeeconf}

\IEEEoverridecommandlockouts  
\usepackage{mathtools}
\usepackage{gensymb}
\usepackage[export]{adjustbox}
\usepackage{graphics} 
\usepackage{epsfig} 
\usepackage{times} 
\usepackage{amsmath} 
\usepackage{tabularx}
\usepackage{amssymb} 
\usepackage{tablefootnote}
\usepackage[table,dvipsnames,svgnames,x11names]{xcolor}
\usepackage{booktabs}
\usepackage{multirow}
\usepackage{multicol}
\usepackage{graphicx}
\usepackage{color, colortbl}
\usepackage[font=small]{caption}
\usepackage{xfrac}
\usepackage{subcaption}     
\usepackage{adjustbox}
\usepackage{lipsum}
\usepackage{layout}

\usepackage[style=ieee, sorting=none, citestyle=numeric-comp, backend=biber]{biblatex}
\usepackage{hyperref}
\hypersetup{
    colorlinks=true,
    citecolor=red,
    linkcolor=red,
}

\usepackage{amsmath,amssymb,amsfonts}

\DeclareMathOperator*{\argmin}{arg\,min}
\usepackage{authblk}
\let\labelindent\relax
\usepackage{enumitem}
\usepackage{breqn}
\usepackage{algpseudocode}
\usepackage[linesnumbered,ruled,lined]{algorithm2e}
\usepackage{array}
\usepackage{makecell}
\usepackage{tabularray}

\title{\LARGE \bf
 \coolname{}: Sparse Open-Set Landmark-based Global Localization \\ for Autonomous Navigation
}

\author{Pranjal Paul$^{*1}$, Vineeth Bhat$^{*1}$, Tejas Salian$^{1}$, Mohammad Omama$^{2}$, \\ Krishna Murthy Jatavallabhula$^{3}$, Naveen Arulselvan$^{4}$, Madhava Krishna$^{1}$
\thanks{$^{*}$ equal contribution}%
\thanks{$^{1}$ Robotics Research Centre, IIIT Hyderabad}%
\thanks{$^{2}$ The University of Texas, Austin}%
\thanks{$^{3}$ MIT}
\thanks{$^{4}$ Ati Motors}
}

\newcommand{\coolname}{\textit{SparseLoc}}\hypersetup{pdfborder={0 0 0}}

\AtEveryCitekey{\bibhyperref{\color{red}}}

\begin{document}

\maketitle
\thispagestyle{empty}
\pagestyle{empty}

\begin{abstract}

Global localization is a critical problem in autonomous navigation, enabling precise positioning without reliance on GPS. Modern techniques often depend on dense LiDAR maps, which, while precise, require extensive storage and computational resources. Alternative approaches have explored sparse maps and learned features, but suffer from poor robustness and generalization. We propose \coolname{}\footnote{Project Page: \href{https://reachpranjal.com/sparseloc}{\texttt{https://reachpranjal.com/sparseloc}}}, a global localization framework that leverages vision-language foundation models to generate sparse, semantic-topometric maps in a zero-shot manner. Our approach combines this representation with Monte Carlo localization enhanced by a novel \emph{late optimization} strategy for improved pose estimation. By constructing compact yet discriminative maps and refining poses through retrospective optimization, \coolname{} overcomes limitations of existing sparse methods, offering a more efficient and robust solution. Our system achieves over 
$5\times$ improvement in localization accuracy compared to existing sparse mapping techniques. Despite utilizing only $1/500^{th}$ of the points used by dense methods, it achieves comparable performance, maintaining average global localization error below $5m$ and $2\degree$ on KITTI. We further demonstrate the practical applicability of our method through cross-sequence localization experiments and downstream navigation tasks.
\vspace{0.2cm}

\end{abstract}
\section{Introduction} \label{introduction}

Mapping and localization at the kilometer scale are central to advancing autonomous driving systems. This involves modeling the surroundings sufficiently to estimate the navigating agent's position within the map, enabling maneuvering throughout urban environments autonomously.

To achieve this, existing industry-led autonomous driving systems meticulously build dense representations of the environment, typically in the form of massive point cloud maps or high-definition maps containing lane-level information \cite{wong2020mapping}. While these approaches provide centimeter-level precision, they impose memory and compute costs that become prohibitive when scaling beyond limited operational domains. Moreover, localizing in such maps generally requires a pose prior to globally estimating the agent's position. Even sophisticated SLAM algorithms \cite{legoloam, lightloam} ultimately depend on tracking continuity or reliable initial pose estimates to recover from what is known as the ``kidnapped robot problem".

Attempting to address these challenges by compressing maps through heavy downsampling, but this merely represents a compromise that weakens accuracy due to the loss of geometrical details inherent to the environment, without fundamentally addressing the scalability challenge. 

Alternatively, oracle-based solutions like GNSS/GPS sensors or navigation maps such as OpenStreetMaps (OSM) are commonly employed \cite{floros2013openstreetslam}. These measurements of location using trilateration methods can become unreliable in urban environments due to multipath effects and occlusions from skyscrapers.

Hence, it is desirable to build a map that is \emph{representative} and can be consumed to perform self-localization - an ability to determine position within the mapped environment without any external assistance like GPS. The idea of a \emph{representative map} follows from the observation that the majority of information in urban environments is concentrated in a small subset of structurally significant points. What truly matters for localization isn't the raw quantity of points, but whether we have the \emph{right points}. However, it can be difficult to simultaneously build a concentrated small subset of structurally significant points and also localize with sufficient precision.

\begin{figure}[!t]
\centering
\includegraphics[width=\linewidth]{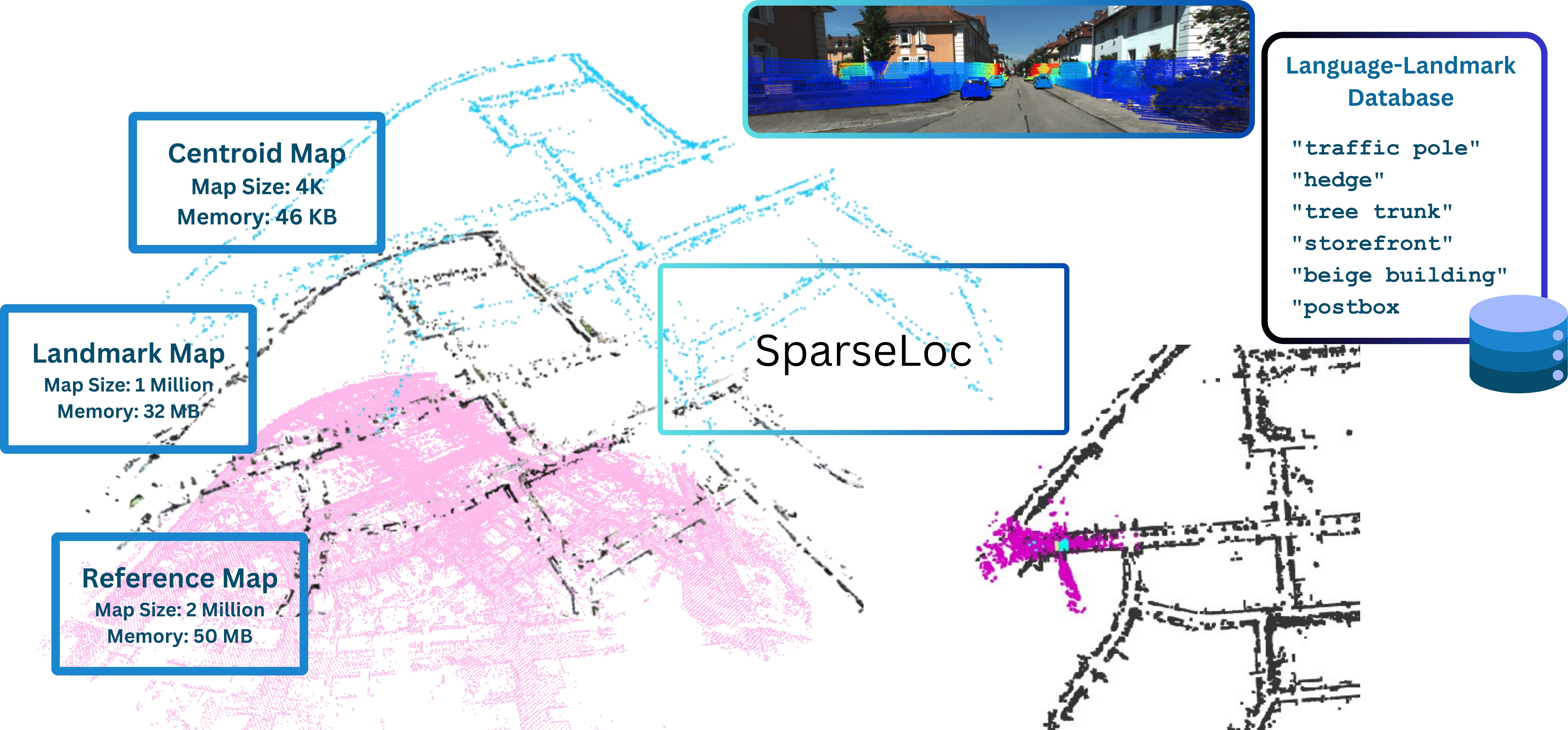}
\caption{\textbf{Introducing \coolname{}}. We introduce \coolname{}, a framework for constructing sparse, language-augmented object-centroid maps using open-set models, a robust data association scheme for accurate Monte Carlo global pose estimation, and a late optimization strategy to minimize localization errors.}  
\label{fig:teaser}
\vspace{-0.4cm}
\end{figure}

Unlike mapping systems that process geometric features in metric space, humans navigate using cognitive spatial maps based on topological cues and distinctive environmental landmarks. While mapping systems see ``just another building or trees" in metric coordinates, humans recognize ``a blue building with adjacent rows of trees". This distinction suggests potential benefits in integrating human-inspired localization principles into SLAM systems to distill environmental representations at the city scale; fundamentally, reimagining what constitutes a ``map point". Our study shows that rather than maintaining an elaborate set of points, we can sparsify the map by retaining a subset of points augmented with foundational features to capture both geometric and semantic nuances. Essentially, we propose a concept of sparsity focusing on reducing the number of map points.

\begin{figure*}[!t]
\centering
\includegraphics[width=\textwidth]{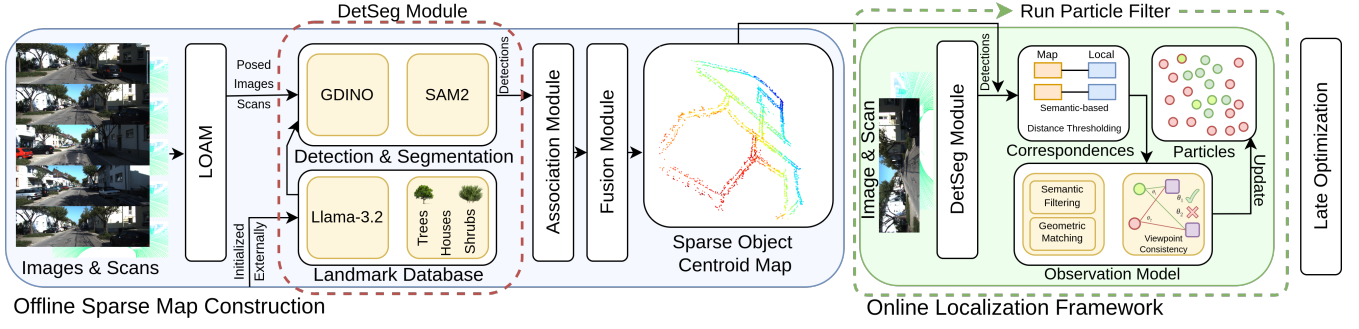}
\caption{\textbf{\coolname{} Architecture.} \small The framework consists of offline semantic-topometric sparse map construction and online global localization. The offline stage builds a compact landmark-based map using open-set models, while the online stage performs localization using Monte Carlo estimation and a novel \textit{Late Optimization} to further refine the pose estimate.}  
\label{fig:architecture}
\end{figure*}

We introduce \textit{SparseLoc}, a global localization framework that utilizes open-vocabulary perception models to construct compact, discriminative maps using landmark cluster centroids for sparse point representation, which can be consumed by downstream localization and navigation modules. Our pipeline comprises of two key stages: an \emph{Offline Sparse Map Construction} and an \emph{Online Global Localization}, as illustrated in Fig. \ref{fig:architecture}.  

Our approach is validated through cross-dataset localization experiments between KITTI-360 Sequence 09 and KITTI-Raw Sequence 07, demonstrating robustness to viewpoint variations where traditional geometric methods generally fail, alongside standard global pose estimation experiments and navigation experiments. 

Our contributions can be summarized as follows:  
\begin{enumerate}  
    \item A framework using vision-language models to create semantically rich sparse maps that enable global localization. (Section \ref{sec:mapping}).  
    \item A Monte Carlo localization \cite{dellaert1999monte} technique with a novel \textit{late optimization} approach showing faster convergence and reduced error. (Section \ref{sec:loc}).  
    \item Empirical results showing improvement by 5 times over existing sparse-map techniques and comparable accuracy to dense-map methods at 1/500$^{th}$ the point density. (Table \ref{tab:dense_ablations_table}, \ref{tab:all_loc}).  
    \item Demonstrated robustness in cross-dataset scenarios with significant viewpoint variations. (Table \ref{tab:cross_loc}).  
    \item System efficacy in CARLA simulation for navigation tasks, evaluated by goal reachability. (Section \ref{sec:carla_res}).  
\end{enumerate}  

\vspace{0.5em}
\noindent\emph{We postulate} that combining classical methods with modern solutions offers the most promising path for autonomous navigation at the city scale, as demonstrated by \coolname{}'s integration of probabilistic localization with foundation models' semantic understanding capabilities.

\section{Related Works} \label{related-works}

\textbf{Global Localization Techniques on LiDAR Maps: }  
For global localization in LiDAR maps, most existing approaches \cite{legoloam, yang2020teaser, lim2022singlequatro} rely on geometric scan matching techniques or graph-theoretic approaches to align the current scan with a pre-registered LiDAR map. More recently, 3D-BBS \cite{aoki20243d} extended branch-and-bound-based 2D scan matching to 3D, achieving state-of-the-art results. However, all these approaches depend on dense LiDAR maps, which are computationally expensive and difficult to store and maintain at scale.

\textbf{Global Localization Techniques using Descriptors: }  
Methods under this category come in the form of place-recognition techniques where a scan descriptor is compared against a database of descriptors \cite{ma2022overlaptransformer, dube2018segmap, uy2018pointnetvlad}. As they retrieve the scan but do not estimate the current pose, there is an additional layer that involves scan matching for pose refinement. Once again, it entails large storage requirements as maps increase in scale.

\textbf{Global Localization Techniques using Sparse Maps: }  
A number of approaches exist that have attempted to represent sparse maps in terms of specific landmarks such as poles \cite{schaefer2019long}, object clusters \cite{peterson2024romanopensetobjectmap}, and OSM descriptors \cite{yan2019global, cho2022openstreetmap}. We leverage object-level information through open-set models. Some fine registration techniques, such as That's My Point \cite{pramatarov2024s}, operate on sparse points using estimates from global localization methods. Additionally, methods that perform fine localization given an initial estimate from an oracle \cite{elhousni2022lidar} have been excluded from discussion, as they assume an external coarse localization source.

\textbf{Vision Language Models: }  
Vision-language models (VLMs) \cite{touvron2023llama, clip} have achieved remarkable success across various domains, including open-set object detection \cite{liu2024grounding} and segmentation \cite{clip}. Recent efforts have explored distilling VLM knowledge into 3D representations \cite{conceptgraphs, jatavallabhula2023conceptfusion}, enabling the creation of open-vocabulary dense 3D maps for downstream tasks. 

Unlike prior works that rely on dense LiDAR scans or handcrafted sparse representations, we introduce a novel approach that utilizes open-world perception models to construct sparse, semantically-meaningful topometric maps for global localization. Our method eliminates the large memory requirement for dense map storage while achieving high localization accuracy, demonstrating the first-ever zero-shot application of VLMs to sparse-map-based LiDAR localization at city-scale.

\section{Methodology} \label{methodology}

\subsection{Sparse Topometric Mapping}
\label{sec:mapping}

\coolname{} builds a lightweight language-augmented instance map of landmark centroids from posed time-synchronized RGB-LiDAR sequences. It uses off-the-shelf vision-language foundation models to identify static open-set landmarks, project the candidate instances to 3D metric space, and associate their cluster centroids across multiple views, constructing a highly sparse representation at city scale that is consumed for the downstream task of localization.

\textbf{Open-Vocabulary Language-Landmark Database:} Our system uses pre-determined open-set labels for both mapping and localization runs. Assuming access to sufficient imagery data that covers the operating region\footnote{While our approach requires an image database which is typically available in common public datasets, in real-world applications, it is a standard practice among existing industry-led autonomous driving systems to conduct initial mapping surveys before deployment.}, we prompt the vision-language model $\mathcal{V}(\cdot)$ over the entire image database to obtain a unique set of text-labels $\mathcal{L} = \{{l_q\}}_{q=1}^{Q}$. This reduces the likelihood of misdetections that might otherwise occur in open-ended detection during localization. Moreover, by curating the database \textit{apriori}, it establishes well-defined anchors- what counts as a landmark in the operating environment. We specifically instruct the model to focus on static objects suitable for localization while excluding dynamic elements such as vehicles and pedestrians\footnote{Complete prompting template is available in the project page}. These descriptions serve as standardized textual identifiers for the landmarks throughout the pipeline. 

\textbf{Landmark 2D Detection and Segmentation:} Given the language-landmark database and an input stream of calibrated \& temporally synchronized RGB-LiDAR observations $\mathcal{Z} = \{(\mathbf{I}_t^{rgb}, \mathbf{I}_t^{lidar}, \mathbf{U}_t)\}_{t=1}^{T}$ (Image, LiDAR Scan, Odometry) with ground-truth poses, we build a centroid map\footnote{Our method can also be followed to distill landmarks from a pre-registered map containing aligned images, point-cloud scans and poses.} $\mathcal{M} = \{(\mathbf{c}_k, l_k, \mathbf{f}_k)\}_{k=1}^{K}$ where each point is characterized by the 3D position of the landmark cluster centroids $\mathbf{c}_k$, a language-aligned feature vector $\mathbf{f}_k$ and the text-label $l_k \in \mathcal{L}$ where $K$ is the total number of \textit{centroids} registered in the final constructed map $\mathcal{M}$.

We process each frame $\mathbf{I}_t^{rgb}$ together with the landmark database $\mathcal{L}$ through a grounding model $\mathcal{G}(\cdot)$ to obtain precise 2D bounding boxes $\mathbf{b}_{t}^{(i)}$ along with their semantic embedding $\mathbf{f}_{t}^{(i)}$ and a confidence score $s_{t}^{(i)}$ directly obtained from the detection model\footnote{Without loss of generality, $\mathcal{G}(\cdot)$ can be followed by dedicated feature extractor $\mathcal{E}(\cdot)$ to obtain semantic embedding $\mathbf{f}_{t}^{(i)} = \mathcal{E}(\mathbf{b}_{t}^{(i)})$ parsing each detected regions.}:
\begin{equation}\label{eq:grounding-model}
{(\mathbf{b}_{t}^{(i)}, \mathbf{f}_{t}^{(i)}, s_{t}^{(i)})}_{i=1}^{N_t} = \mathcal{G}(\mathbf{I}_t^{\text{rgb}}, \{l_{q}\}_{q=1}^{Q})
\end{equation}

where $N_t$ is the number of detected landmark instances at frame $t$. Further, when processing the calibrated input frame $(\mathbf{I}_t^{rgb}, \mathbf{I}_t^{lidar})$, we query a class-agnostic segmentation model $\mathcal{S(\cdot)}$ for each detected region to obtain a pixel-level mask: 
\begin{equation} \label{eq:seg-model}
\mathbf{m}_{t}^{(i)} = \mathcal{S}(\mathbf{I}_t^{rgb}, \mathbf{b}_{t}^{(i)}) 
\end{equation}

Each masked 2D instance is projected into 3D space using the calibration matrix, refined through DBSCAN clustering, and transformed to the map frame using the corresponding odometry pose $\mathbf{U}_t$. We then compute the centroid ${}^{\text{map}}\mathbf{c}_{t}^{(i)}$ for each $i^{th}$ cluster as the resultant instance observations $\mathbf{o}_{t}^{(i)} = \langle {}^{\text{map}}\mathbf{c}_{t}^{(i)}, \mathbf{f}_{t}^{(i)} \rangle$. Note that, for cases where the overlapping observations for different labels map to the same physical object, we tag the one with the highest confidence score $s_t$ to the resulting cluster\footnote{For the semantically similar observations, their feature vector can also be averaged out.}. This ensures that each point in the map has a semantically accurate description.

\textbf{Instance Association and Fusion:} Our association strategy mostly follows from \cite{conceptgraphs} and has been overviewed in Fig. \ref{fig:architecture} as the association and fusion modules. As we detect new landmark instances $\mathbf{o}_{t}^{(i)}$, we evaluate their correspondence scores with all the existing instances $\mathbf{o}_{t-1}^{(k)}$ registered so far in the map by computing: (1) the geometric similarity $\phi_{geo}(i, k)$ which is the spatial proximity between instance clusters based on a nearest-neighbor ratio within a certain threshold, and (2) the semantic affinity measure $\phi_{sem}(i, k)$, a normalized cosine similarity between their feature vectors: $\phi_{\text{sem}}(i, k) = \mathbf{f}_{i}^T \mathbf{f}_{k}/{2} + 1/2$. The overall similarity score $\phi(i, k)$ is the sum of both: $\phi(i, k) = \phi_{sem}(i, k) + \phi_{geo}(i, k)$. An instance association is established if $\max_{k} \phi(i, k) > \delta_{sim}$. In this case, we update the matched instance with the new observation by updating the point cloud, updating the feature vector through confidence-weighted averaging, and incrementing the observation count. If no suitable match is found, we initialize a new instance in the map.

\subsection{Multimodal Localization}
\label{sec:loc}

Localization in large-scale environments presents significant challenges, particularly within the proposed framework due to: (1) the structural sparsity of the global map representation and (2) mean shifts in the landmark point cloud across scans, causing positional instability that prevents consistent 3D landmark correspondence. These factors hinder precise global alignment using a single scan.
To address these challenges, we employ particle-filter-based localization, a recursive Bayesian state estimation method well-suited for incrementally refining pose estimates and achieving global convergence from multiple hypotheses as motion observations accumulate. However, robust data association is critical for consistent pose refinement, forming the core of our approach.

\textbf{Problem Setup:} Given a sequence of multimodal sensor observations $\mathcal{Z}_t = \{(\mathbf{I}_t^{rgb}, \mathbf{I}_t^{lidar})\}_{t=1}^T$, a relative odometry source $\mathbf{U}_t = \{\mathbf{u_t}\}_{t=1}^{T}$ from some standard SLAM pipeline and our previously built semantically-augmented reference map $\mathcal{M}$ together with language tags $\mathcal{L}$, we want to estimate a posterior distribution $p(\mathbf{T}_t | \mathcal{Z}_{1:t}, \mathbf{U}_{1:t}, \mathcal{M}, \mathcal{L})$ of the robot's state $\mathbf{T}_t \in SE(3)$ by approximating the distribution using $N$ weighted pose hypotheses (particles) $\{\mathbf{T}_t^{(i)}, w_t^{(i)}\}_{i=1}^N$ where $w_t^{(i)}$ denotes the importance weight of the $i^{th}$ particle:
\begin{equation}
w_t^{(i)} \propto p(\mathcal{Z}_t \;|\; \mathbf{T}_t^{(i)}, \mathcal{M}, \mathcal{L}) \cdot p(\mathbf{T}_t^{(i)} \;|\; \mathbf{T}_{t-1}^{(i)}, \mathbf{u}_t)
\end{equation}

Our objective is to establish a multimodal observation likelihood function that returns optimal probabilistic correspondences between observed landmark centroids and the map landmarks.

\textbf{Prediction Step:} At initialization, we spawn $N$ particles randomly across the map $\mathcal{M}$. Each particle $\mathbf{T}_t^{(i)}$ is dead-reckoned using the relative odometry measurements, perturbed with Gaussian noise to account for odometry uncertainty. While this provides a reasonable pose prior, it accumulates drift over time.

\textbf{Update Step:} To correct these accumulated errors, we define a refinement step that associates the observed landmark instances with the map context. Each incoming data is processed identically to the mapping phase to detect and segment open-set landmarks using equation (\ref{eq:grounding-model}) and equation (\ref{eq:seg-model}) followed by extracting corresponding 3D point clusters. For each particle pose hypothesis $\mathbf{T}_t^{(i)}$, we compute the observation likelihood by comparing observed landmarks with map landmarks.

\textbf{Data Association and Weighting Mechanism:} We integrate both semantic and geometric landmark information for robust data association. Each particle represents a hypothesized robot pose, and if correct, the observed landmarks should align with mapped landmarks when transformed accordingly. This alignment is quantified by computing a weight contribution for each particle-landmark pair as follows:

\begin{enumerate}
    \item \textbf{Semantic Filtering:} For each detected instance centroid $\langle {}^{\text{local}}\mathbf{c}_{t}^{(j)}, \mathbf{f}_{t}^{(j)} \rangle$, we first extract candidate map centroids using cosine similarity $\phi_{sem}(j, k) > \delta_{sem}$ and transform them to particle's frame $
{}^{local}\hat{\mathbf{c}}_{k}^{(i)} = \mathbf{T}_t^{(i)^{-1}} \cdot {}^{map}\mathbf{c}_{k}
$ where $\mathbf{c}_k$ is the candidate centroid filtered from the map (abstracted as \emph{``DetSeg"} module in Fig. \ref{fig:architecture}).

    \item \textbf{Geometric Matching:} The geometric compatibility is then computed based on the Euclidean distance between them and converted into a distance score using the exponential decay function: $\phi_{dist}(j, k, i) = \exp\left(-\frac{e_d}{\alpha_{dist}}\right)$ where $e_d = ||{}^{local}\hat{\mathbf{c}}_{k}^{(i)} - {}^{local}\mathbf{c}_{t}^{(j)}||$ and $\alpha_{dist}$ is a scaling parameter. This creates a soft matching score that decreases with distance.

    \item \textbf{Viewpoint Consistency:} Euclidean distance alone is insufficient as it is viewpoint-agnostic. A landmark observed from a specific viewing direction in the sensor frame should, under a correct pose hypothesis, appear from a consistent direction in the particle frame. To handle this, viewing angle consistency is assessed by computing the bearing angles of each landmark relative to all three coordinate axes (x, y, z) and comparing the observed directions with the expected directions under the particle hypothesis:
    \begin{equation}
    \phi_{\text{angle}}(j, k, i) = 1/3\sum_{\alpha \in \{x,y,z\}} \cos(\Delta\theta_{\alpha}) + 1/2 \cdot \beta \quad
    \end{equation}
    
    where $\beta = 1/(N \cdot \lambda)$ with scaling factor $\lambda$ and total number of particles $N$. $\Delta\theta_{\alpha}$ is the angular difference between the observed landmark direction and the expected direction under the particle hypothesis when projected onto each coordinate axis. This accounts for full 3D viewing direction consistency across all rotational degrees of freedom.

\item \textbf{Overall Particle Weight:} Particle weights are updated exclusively based on current landmark observations, independent of prior weights. This enables the state estimator to adjust to new measurements rapidly without bias from past estimates. Each weight is determined by summing the highest-scoring landmark correspondences in the current frame. 
$w_t^{(i)} \propto \sum_{j=1}^{N_L} \max_k \phi_{total}$
where $\phi_{total}(j, k, i) = \phi_{sem} \cdot \phi_{dist} + \phi_{angle}$ which are then normalized using softmax with temperature $\alpha_{softmax}$.

\end{enumerate}

\textbf{Resampling and Pose Estimation:} We periodically resample particles with replacement to concentrate in high-probability regions. The final pose estimate is then computed as the weighted average of all particle poses. For translation components, we use standard weighted averaging:

\begin{equation}
\mathbf{t}_t = \sum_{i=1}^{N} w_t^{(i){\gamma}} \cdot \mathbf{t}_t^{(i)}
\end{equation}

where $\gamma$ is a weight exponent parameter to control the influence of higher-weight particles. For rotation, we extract Euler angles from each SE(3) transformation and apply circular averaging separately for each angle:
\begin{equation}    
\theta_{t,\alpha} = \arctan2\left(\sum_{i=1}^N w_t^{(i)\gamma} \sin(\theta_{t,\alpha}^{(i)}), \sum_{i=1}^N w_t^{(i)\gamma} \cos(\theta_{t,\alpha}^{(i)})\right)
\end{equation}
where $\alpha \in \{\text{roll, pitch, yaw}\}$.

This prevents discontinuities that would occur with direct averaging of angles. For example, naively averaging $359^\circ$ and $1^\circ$ would yield $180^\circ$, while circular averaging correctly produces $0^\circ$, respecting the true shortest angular distance between the values. We apply this circular averaging independently to each of the three Euler angles extracted from the SE(3) transformations.

\textbf{Late Optimization:} Once the particle filter converges to a coarse estimate, we perform pose refinement through a novel process that shares principles with pose-graph optimization. The key insight is to anchor the converged pose and retroactively improve the correspondences for previous observations within a history window $\mathcal{H}$, which will be leveraged to perform global optimization. 

Given the converged particle pose $\mathbf{T}_t$ at the current time $t$, we compute the refined pose for each previous timestamp $t^- < t$: $\mathbf{T}_{t^-} = \mathbf{T}_t \times (\mathbf{T}^{rel}_{t^- \rightarrow t})^{-1} $ where $\mathbf{T}^{rel}_{t^- \rightarrow t}$ represents the cumulative relative transformation from time $t-\mathcal{H}$ to $t$ derived from odometry measurements. This relation can be expanded as the product of sequential relative transformations that were used in the predictive (dead-reckoning) steps:
\begin{equation}
\mathbf{T}^{rel}_{t^- \rightarrow t} = \mathbf{T}^{rel}_{t^-+1} \cdot \mathbf{T}^{rel}_{t^-+2} \cdot ... \cdot \mathbf{T}^{rel}_t
\end{equation}

Using these refined poses, we recompute landmark-to-map correspondences\footnote{Direct ICP alignment was unstable due to sparse map density and the requirement for hard correspondences. We evaluated two alternatives: RANSAC with the Orthogonal Procrustes solution and an NLLS optimization over SE(3) with Huber loss yielding equivalent results.} for each previous frame, resulting in correspondence sets $\mathcal{C}_t$, through a comprehensive matching process. This involves: (1) detecting landmark instances from sensor observations, (2) computing feature vectors for each detected instance, (3) matching these instances with candidate landmarks using cosine similarity of feature vectors, (4) filtering matches based on distance thresholds and positional constraints, and (5) constructing source-target centroid pairs for optimization. 

We then perform a global optimization that considers $\mathcal{C}_t$ to refine the particle pose. 

\begin{equation}\label{Refine}
\mathbf{T}^* = \argmin_{\mathbf{T}} \sum_{i \in \mathcal{H}} \sum_{(j,k) \in \mathcal{C}_t} ||\mathbf{T} \cdot {}^{local}\mathbf{c}_{j}^{(i)} - {}^{map}\mathbf{c}_{k}||^2
\end{equation}

where $\mathbf{T}$ is the optimization variable. This global optimization finds a single transformation $\mathbf{T}^*$ that best aligns all the previous landmarks observations in the history window $\mathcal{H}$ to the map structure. 

\subsection{Implementation Details}

\coolname{} offers flexibility in model selection, supporting any open-world detection and segmentation model, large vision-language models for landmark database construction, and any odometry pipeline. For our experiments, we use Llama-3.2-Vision \cite{touvron2023llama} for creating language-landmark database $\mathcal{L}$, Grounding-DINO \cite{liu2024grounding} for detection and feature extraction $\mathcal{G}(\cdot)$ and SAM2 \cite{ravi2024sam} as segmentation model $\mathcal{S}(\cdot)$. We use LeGO-LOAM \cite{legoloam} for odometry estimation and reference map generation, obtaining temporally synchronized RGB-LiDAR scans.

For the mapping phase, we choose a similarity score threshold of $\delta_{sim} = 0.7$ for instance association. A new instance is created if $\phi_{sem}$ falls below this threshold. For localization, we initialize $N = 1000$ particles randomly and uniformly distributed over the entire map. In the data association step, we set the semantic similarity threshold to $\delta_{sem} = 0.9$, the geometric matching scale factor $\alpha_{dist} = 1$, and the viewpoint consistency coefficient $\alpha = 10^{-3}$. The temperature $\alpha_{softmax}$ is set to $0.5$ for normalized particle weight computation, and the weight exponent parameter $\gamma$ is set to $1$. For late optimization, we use a history window of $\mathcal{H} = 10$ frames.

\section{Results} \label{results}

\subsection{Experimental Setup}  

We evaluate our approach on the KITTI-Odometry dataset \cite{kittigeiger2013vision}, focusing on Sequences $00, 05$, and $07$. These sequences provide kilometer-scale urban runs and multiple revisits suitable for evaluating our localization approach. Additionally, we perform cross-dataset validation by aligning segments of Sequence $09$ from the KITTI-360 dataset \cite{kitti360liao2022kitti} onto the sparse map constructed from Sequence $07$ of KITTI-Odometry. Table \ref{tab:map_size} compares the map sizes between our method and LOAM \cite{legoloam} across different KITTI-Odometry sequences.

\begin{table}[!h]
\centering
\footnotesize
\renewcommand{\arraystretch}{1.5}
\setlength{\tabcolsep}{3pt}
\begin{tabular*}{\columnwidth}{@{\extracolsep{\fill}}>{\centering\arraybackslash}p{2.5cm}|>{\centering\arraybackslash}p{1.5cm}|>{\centering\arraybackslash}p{1.5cm}|>{\centering\arraybackslash}p{1.5cm}@{}}
\toprule
 & \textbf{Seq. 00} & \textbf{Seq. 05} & \textbf{Seq. 07} \\ 
\midrule
\textbf{Map Run Length} & 3,714 m & 2,223 m & 695 m\\ 
\midrule
\multicolumn{4}{c}{\textbf{Map Size (No. of Registered Points)}} \\
\midrule
LOAM & 1,386,822 & 1,899,437 & 437,149 \\
\addlinespace[1.5pt]\hline\addlinespace[1.5pt] 
\textit{\textbf{Ours}} & \textbf{3,828} & \textbf{3,350} & \textbf{952} \\
\bottomrule
\end{tabular*}
\caption{\textbf{Map Size Comparison.} Point Cloud reduction achieved by \coolname{}. Our map requires only thousands of points versus millions needed by LOAM \cite{legoloam}.}
\label{tab:map_size}
\end{table}

We assess localization performance through average error metrics- Average Translation Error (ATE) and Average Rotation Error (ARE), and success rate thresholds. The former measures overall accuracy across the entire run, while the latter indicates how consistently the system meets predefined precision criteria. 

\begin{table*}[!ht]
\centering
\footnotesize
\renewcommand{\arraystretch}{1.5}
\setlength{\tabcolsep}{3pt}
\begin{tabular*}{\textwidth}{@{\extracolsep{\fill}}c|ccc|ccc|ccc|ccc|ccc|ccc@{}}
\toprule
\multirow{3}{*}{\textbf{Method}} 
  & \multicolumn{6}{c|}{\textbf{Seq. 00}} 
  & \multicolumn{6}{c|}{\textbf{Seq. 05}} 
  & \multicolumn{6}{c}{\textbf{Seq. 07}} \\
\cmidrule{2-19}
& \multicolumn{3}{c|}{$(4m, 3\degree)$} & \multicolumn{3}{c|}{$(10m, 5\degree)$}
& \multicolumn{3}{c|}{$(4m, 3\degree)$} & \multicolumn{3}{c|}{$(10m, 5\degree)$}
& \multicolumn{3}{c|}{$(4m, 3\degree)$} & \multicolumn{3}{c}{$(10m, 5\degree)$} \\
\cmidrule{2-19}
& $t_{avg}$ & $r_{avg}$ & $SR$
& $t_{avg}$ & $r_{avg}$ & $SR$
& $t_{avg}$ & $r_{avg}$ & $SR$
& $t_{avg}$ & $r_{avg}$ & $SR$
& $t_{avg}$ & $r_{avg}$ & $SR$
& $t_{avg}$ & $r_{avg}$ & $SR$ \\
\midrule
KISS-Matcher & \textbf{0.321} & 1.382 & 33.54 
  & \textbf{0.436} & 1.912 & 42.41 
  & \textbf{0.299} & 1.166 & 44.57 
  & \textbf{0.340} & \emph{1.569} & 53.26 
  & \textbf{0.351} & 1.043 & 77.42 
  & \textbf{0.353} & 1.140 & 80.65 \\
\midrule
3D-BBS & \emph{0.714} & \textbf{0.024} & \textbf{86.16} 
  & \emph{0.714} & \textbf{0.024} & \emph{86.16} 
  & \emph{0.814} & \textbf{0.019} & \textbf{75.00} 
  & \emph{0.814} & \textbf{0.019} & \emph{75.00} 
  & 0.759 & \textbf{0.025} & \emph{100.0} 
  & 0.759 & \textbf{0.025} & \emph{100.0} \\
\midrule
\textbf{\textit{Ours}} & 2.103 & \emph{1.294} & \emph{50.67} 
  & 4.054 & \emph{1.451} & \textbf{99.33} 
  & 1.293 & \emph{1.066} & \emph{48.31} 
  & 3.150 & \emph{1.451} & \textbf{99.33} 
  & \emph{0.723} & \emph{0.417} & \textbf{100.0} 
  & \emph{0.723} & \emph{0.417} & \textbf{100.0} \\
\bottomrule
\end{tabular*}
\caption{\textbf{Quantitative Comparison of Global Localization Performance.} Reports pose accuracy using standard evaluation metrics- Average Translation Error (ATE) ($t_{avg}$, in meters), Average Rotation Error (ARE) ($r_{avg}$, in degrees), and Success Rate ($SR\%$) at two different accuracy thresholds: $(4m, 3°)$ and $(10m, 5°)$. Best result shown in \textbf{bold} font, while the second-best is \emph{italicized}.}
\label{tab:dense_ablations_table}
\end{table*}

\subsection{Global Localization Results} \label{sec:global_loc_res} 

\subsubsection{\textbf{Comparison with Dense LiDAR Map-Based Localization}} We compare our approach against two leading dense registration methods: 3D-BBS~\cite{aoki20243d}, which has positioned itself as SOTA in global localization through favorable comparisons with popular methods \cite{yang2020teaser, lim2022singlequatro}, and KISS-Matcher~\cite{lim2024kiss} for its strong performance in scan-to-scan and scan-to-map registration. 3D-BBS is a search-based method that systematically explores $6$-DoF pose space via branch-and-bound, whereas KISS-Matcher is a descriptor-based method using enhanced FPFH features with graph-theoretic outlier rejection.

Our results across Tables \ref{tab:dense_ablations_table} and \ref{tab:all_loc} demonstrate consistent performance advantages. We outperform KISS-Matcher in success rates at the stricter $(4m, 3\degree)$ threshold and surpass both dense baselines at the relaxed $(10m, 5\degree)$ threshold. Notably, with nearly $500\times$ smaller map representation, we achieve over $99\%$ success rates for Sequences $00$ and $05$, and $100\%$ success rate for Sequence $07$. Table \ref{tab:all_loc} further demonstrates our efficiency in terms of average errors over the entire run, where our sparse approach achieves significantly lower translation errors than KISS-Matcher ($4.064m$ vs $107.0m$ for Sequence $00$) while maintaining competitive rotation accuracy.

While dense methods achieve high precision, their lower success rates reveal challenges with perceptual aliasing in repetitive urban structures. By exploiting both open-set foundation features and multiple hypothesis evaluation through particle filtering, our approach provides better coverage of the solution space, effectively disambiguating similar-looking environments to achieve consistently higher success rates.

\vspace{0.3cm}
\subsubsection{\textbf{Comparison with Sparse Map-Based Localization}} 
Table~\ref{tab:all_loc} compares our method against \cite{yan2019global} using evaluation data from \cite{elhousni2022lidar}. Our approach localizes within $20-40$ KITTI frames ($\sim 20m$ run) compared to the baseline requirement of $500$ frames, while maintaining similar map sizes ($4.0K$ vs. $4.5K$ landmarks). The translation error comparison shows consistent improvements: $4.064m$ vs. $20m$ (Sequence $00$), $5.456m$ vs. $25.00m$ (Sequence $05$), and $0.723m$ vs.
$25.00m$ (Sequence $07$). This corresponds to an average $80\%$ error reduction. Additionally, our method estimates full $6$-DoF poses while the OSM-based method \cite{yan2019global} provides only translation estimates.

\begin{table}[!h]
\centering
\footnotesize
\renewcommand{\arraystretch}{1.5}
\setlength{\tabcolsep}{4pt}
\begin{tabular}{@{}c|c|cc|cc|cc@{}}
\toprule
\multirow{2}{*}{\textbf{Type}} & \multirow{2}{*}{\textbf{Methods}} & \multicolumn{2}{c|}{\textbf{Seq. 00}} & \multicolumn{2}{c|}{\textbf{Seq. 05}} & \multicolumn{2}{c}{\textbf{Seq. 07}} \\
\cmidrule(lr){3-4} \cmidrule(lr){5-6} \cmidrule(l){7-8}
 & & $t_{avg}$ & $r_{avg}$ & $t_{avg}$ & $r_{avg}$ & $t_{avg}$ & $r_{avg}$ \\
\midrule
\multirow{2}{*}{\rotatebox[origin=c]{90}{\small Dense}} 
 & KISS-Matcher \cite{lim2024kiss} & 107.0 & 59.21 & 76.88 & 38.21 & 5.640 & 11.35 \\
\addlinespace[2pt]
 & 3D-BBS \cite{aoki20243d} & 27.71 & \textbf{0.335} & 33.18 & \textbf{0.503} & 0.759 & \textbf{0.025} \\
\midrule
\midrule
\addlinespace[2pt]
\multirow{2}{*}{\rotatebox[origin=c]{90}{\small Sparse}} 
 & OSM 4-bit \cite{yan2019global} & 20.00 & -- & 25.00 & -- & 25.00 & -- \\
\addlinespace[2pt]
 & \textit{\textbf{Ours}} & \textbf{4.064} & 1.481 & \textbf{5.456} & 2.222 & \textbf{0.723} & 0.417 \\
\bottomrule
\end{tabular}
\caption{\textbf{Comparison of Registration Methods.} Shows ATE $(m)$ and ARE $(\degree)$. Best results in \textbf{bold}.}
\label{tab:all_loc}
\end{table}
\vspace{-0.1cm}

These results indicate that sparse landmark-based localization can approach dense method accuracy levels while maintaining computational efficiency. Our approach uses $250-500$ times fewer map points than the baselines, which are dense registration methods, yet achieves competitive performance across varied environments.

\vspace{0.3cm}
\subsubsection{\textbf{Cross-dataset Localization}} Cross-dataset localization presents a significant challenge, as localization is attempted across sequences captured at different times, often years apart. While these sequences may share substantial observation overlap, the set of sparse landmarks common to both becomes even sparser, exacerbating the difficulty of matching across substantial viewpoint and environmental variations.
Our method successfully localizes observations from Sequence $09$ of KITTI-360 within a sparse map generated from Sequence $07$ of KITTI-Odometry, despite the two sequences being captured approximately two years apart. This demonstrates the robustness of our approach to temporal and viewpoint changes. 

To quantitatively assess cross-sequence localization accuracy, we perform coarse registration between Sequences $09$ and $07$, enabling transformation of estimated poses from Sequence $07$'s coordinate frame to Sequence $09$'s frame. Table \ref{tab:cross_loc} presents the evaluation results, demonstrating that over $80\%$ of observations from sequence $09$ are successfully localized within $10m$ translation and $5\degree$ rotation thresholds when using the sparse map constructed from Sequence $07$.

\begin{table}[!h]
\centering
\footnotesize
\renewcommand{\arraystretch}{1.5}
\setlength{\tabcolsep}{1pt}
\begin{tabular*}{\columnwidth}{@{\extracolsep{\fill}}>{\centering\arraybackslash}p{2.2cm}|>{\centering\arraybackslash}p{2cm}|>{\centering\arraybackslash}p{2cm}|>{\centering\arraybackslash}p{2.2cm}@{}}
\toprule
\textbf{Threshold} & \textbf{$t_{\text{avg}}$ $(m)$} & \textbf{$r_{\text{avg}}$ $(\degree)$} & \textbf{SR $(\%)$} \\
\midrule
$4m, 3\degree$ & 1.280 & 1.191 & 57.69 \\
\addlinespace[1.5pt]\hline\addlinespace[1.5pt]
$10m, 5\degree$ & 2.065 & 1.672 & 80.77 \\
\addlinespace[1.5pt]\hline\addlinespace[1.5pt]
$15m, 7\degree$ & 2.463 & 1.636 & 84.62 \\
\bottomrule
\end{tabular*}
\caption{\textbf{Cross-Dataset Localization}: Registration of scans from KITTI-360 Sequence 09 on a sparse map generated from KITTI-Odometry Sequence 07. Reports ATE $(m)$, ARE $(\degree)$, and Success Rate for different thresholds over the entire run.}
\label{tab:cross_loc}
\end{table}
\vspace{-0.3cm}

\subsection{Effect of the Late Optimization Schedule}

We initially attempted standard ICP refinement given our reasonable coarse registration and established correspondences. However, the insufficient point density within the optimization window caused convergence failures, motivating our custom Late Optimization approach designed specifically for sparse semantic features. The substantial improvements in Table \ref{tab:late_opt_ablation} validate this design choice over traditional geometric refinement methods. 

We compare our coarse particle filter registration against the full pipeline that includes pose refinement using Late Optimization. The results reveal distinct performance patterns: the modest but consistent gains in Sequences $00$ and $05$ ($16.3\%$ and $16.4\%$) suggest that in longer trajectories with moderate landmark density, Late Optimization primarily corrects accumulated drift. In contrast, the $57\%$ improvement in Sequence $07$ indicates that shorter trajectories with higher landmark density benefit disproportionately from retrospective analysis. 

\begin{table}[h]
\centering
\footnotesize
\renewcommand{\arraystretch}{1.5}
\setlength{\tabcolsep}{3pt}
\begin{tabular*}{\columnwidth}{@{\extracolsep{\fill}}>{\centering\arraybackslash}p{2.5cm}|>{\centering\arraybackslash}p{1.8cm}|>{\centering\arraybackslash}p{1.8cm}|>{\centering\arraybackslash}p{1.8cm}@{}}
\toprule
\textbf{Method} & \textbf{Seq. 00} & \textbf{Seq. 05} & \textbf{Seq. 07} \\
\midrule
Particle Filter Only & 4.85 & 6.70 & 1.80 \\ 
\addlinespace[1.5pt]\hline\addlinespace[1.5pt]
+ Late Optimization & \textbf{4.06} & \textbf{5.60} & \textbf{0.80} \\
\midrule
\addlinespace[1.5pt]\hline\addlinespace[1.5pt]
\textit{\%age Improvement} & \emph{16.3\%} & \emph{16.4\%} & \emph{55.6\%} \\
\bottomrule
\end{tabular*}
\caption{\textbf{Ablation Study of Late-Optimization.} Localization accuracy improvement with our novel pose refinement operation. Translation errors shown in meters, with percentage improvements in \emph{italics}. Best results highlighted in \textbf{bold}.}
\label{tab:late_opt_ablation}
\end{table}

This performance differential suggests that our refinement strategy excels when sufficient observational context is available, enabling Late Optimization to resolve ambiguities that may constrain the initial particle filter estimates.

\subsection{Navigation with Sparse Maps} \label{sec:carla_res}

\begin{figure}[!h]
    \centering
    \includegraphics[width=\columnwidth]{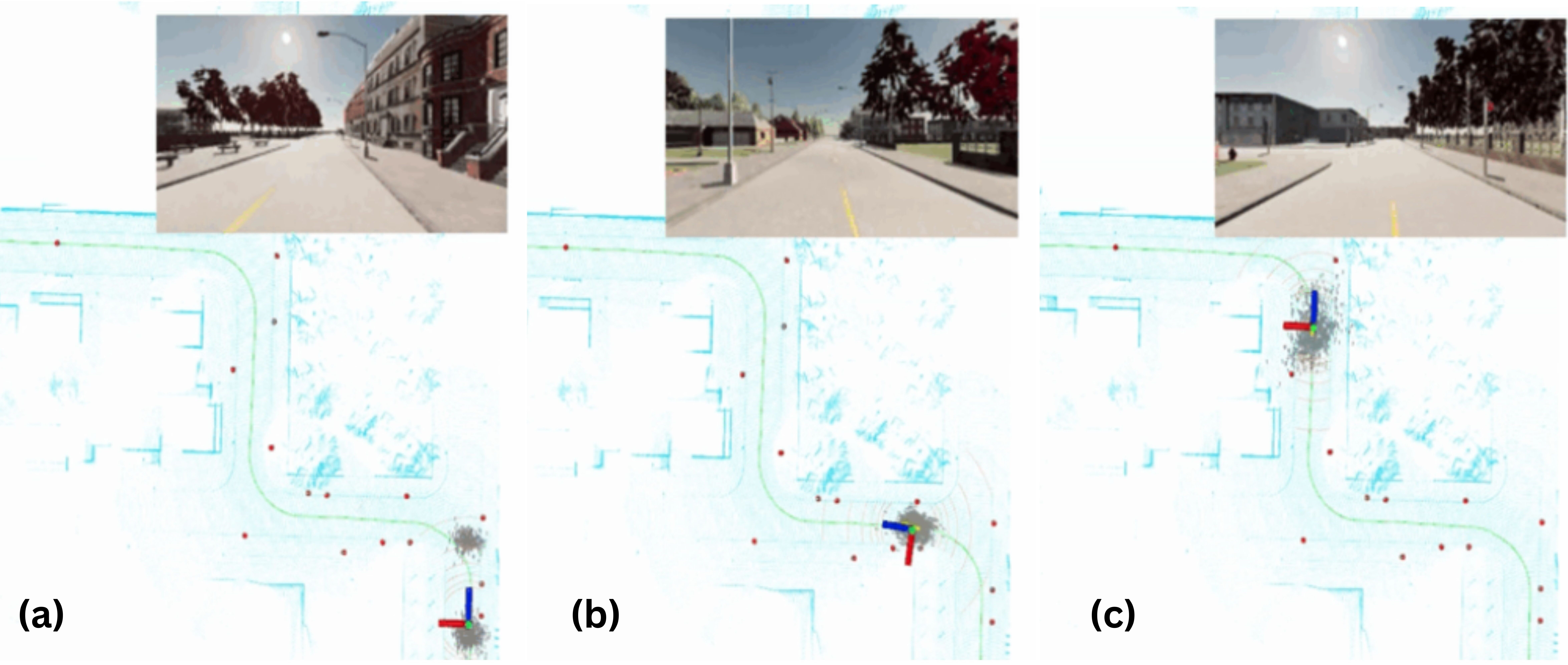}
    \caption{\textbf{CARLA Navigation Runs}. \small Navigation sequence on Town 01 from left to right. The \textcolor{red}{red} points are the mapped landmarks.The grey particle cloud and its mean show the filter converging from a multi-modal to unimodal distribution, maintaining convergence even through the first turn. Point Cloud (in \textcolor[RGB]{0,173,239}{blue}) is shown only for visualization.}
    \label{fig:navigation}
    \vspace{-1em}
\end{figure}

In this section, we demonstrate the effectiveness of our sparsely constructed open-set feature augmented map for the downstream task of navigation. Fig. \ref{fig:navigation} illustrates point-to-point navigation run on CARLA Town 01. The red points represent registered landmark centroids utilized during relocalization. We evaluate the navigation performance based on goal reachability by measuring the distance between the vehicle's final position and the goal location.

\begin{table}[!h]
\centering
\footnotesize
\renewcommand{\arraystretch}{1.5}
\setlength{\tabcolsep}{2pt}
\begin{tabular*}{\columnwidth}{@{\extracolsep{\fill}}>{\centering\arraybackslash}p{2.5cm}|>{\centering\arraybackslash}p{2cm}|>{\centering\arraybackslash}p{1.8cm}|>{\centering\arraybackslash}p{1.8cm}@{}}
\toprule
\makecell{\textbf{Map Density}} & \makecell{\textbf{Map Length} \\ \textbf{$(m)$}} & \makecell{\textbf{Town 01} \\ \textbf{$(m)$}} & \makecell{\textbf{Town 02} \\ \textbf{$(m)$}} \\
\midrule
300 landmarks & 1,500 & 1.55 & 1.63 \\
\bottomrule
\end{tabular*}
\caption{\textbf{Goal Reaching Error.} Closed-loop navigation on CARLA towns with only 300 map points registered using the same landmark labels on both the towns.}
\label{tab:closedloop}
\end{table}

During navigation, the vehicle encounters viewpoint changes at turns and intersections that may cause perceptual aliasing. However, due to the multimodal nature of our particle filter approach, similar-looking landmarks quickly disambiguate as the vehicle continues to move and converge faster to the correct location estimate. As shown in Table \ref{tab:closedloop}, we consistently achieved goal reaching within a proximity of $\sim2m$, showcasing the efficacy of our sparse map for navigation.

\subsection{Role of Foundation Model and Sparsity Analysis}

\begin{figure}[!h]
    \centering
    \includegraphics[width=0.95\linewidth]{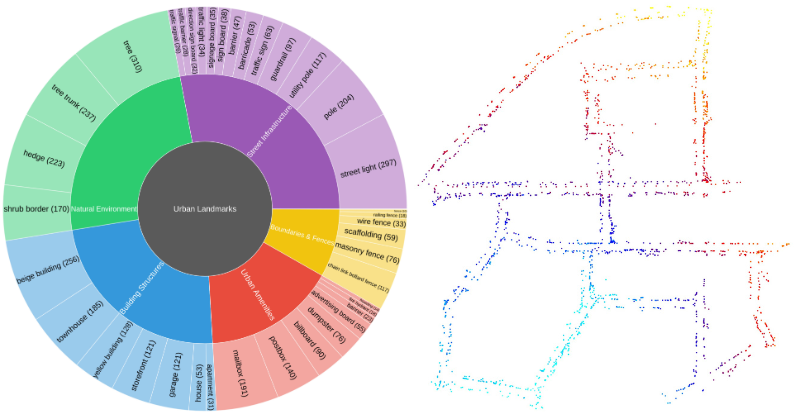}
    \caption{\textbf{SparseLoc Database.} \small \textbf{Left:} shows the distribution of the landmarks registered. \textbf{Right:} Shows the registered landmark centroid map of KITTI Sequence 00. They are sparse but reasonably distributed in urban scenes throughout the scale of the map. (Best viewed at $5\times$ zoom)}
    \label{fig:lm_distrb}
\end{figure}

Foundation models, trained on web-scale data, excel at general semantic concepts. Our framework taps into the capabilities of such Open-World Perception models for localization through intuitive, zero-shot prompting to identify landmarks static to the scene. Remarkably, a single prompt-generated database (Fig. \ref{fig:lm_distrb}) from KITTI Sequence 00 enabled robust cross-sequence and cross-dataset localization, handling significant viewpoint/appearance variations without retraining. This demonstrates VLMs' ability to identify universally useful, distinctive landmarks at the city scale.

While landmark sparsity causes visual ambiguity (e.g., frequent trees dominate rare objects), our framework maintains high accuracy by integrating VLM semantics with particle filtering. Our work highlights this synergy between classical methods and modern AI, blending their strengths to create a robust, practical system for autonomous navigation.

\section{Conclusion} \label{conclusion}

We presented \coolname{}, a global localization framework that integrates vision-language foundation models with Monte Carlo localization to generate and operate within highly sparse semantic maps. Our zero-shot approach achieves effective localization using only a fraction of typical map density, dramatically reducing the computational and storage overhead of dense LiDAR-based systems.

However, our approach is bounded by the computational limitations of the detection and segmentation modules, preventing real-time operation. Additionally, at stricter thresholds, the severe map sparsity limits localization precision. Future work on optimal landmark densities and real-time vision-language models could address these constraints. These results validate our approach as a scalable, memory-efficient solution for autonomous navigation systems.

\newpage
\printbibliography

\end{document}